\documentclass{article}

\usepackage[final, nonatbib]{neurips_2020}

\usepackage[utf8]{inputenc} 
\usepackage[T1]{fontenc}    
\usepackage{hyperref}       
\usepackage{url}            
\usepackage{booktabs}       
\usepackage{amsfonts}       
\usepackage{nicefrac}       
\usepackage{microtype}      
\usepackage{graphicx}
\usepackage{ amssymb }

\title{Biomechanical modelling of brain atrophy through deep learning}

\usepackage{authblk}

\author[1,*]{\textbf{Mariana da Silva}}
\author[2]{\textbf{Kara Garcia}}
\author[1,3,4]{\textbf{Carole H. Sudre}}
\author[1]{\textbf{Cher Bass}}
\author[1]{\textbf{M. Jorge Cardoso}}
\author[1]{\textbf{Emma Robinson\vspace{-5pt}}}

\affil[1]{School of Biomedical Engineering and Imaging Sciences, King's College London}
\affil[2]{Department of Radiology and Imaging Sciences, School of Medicine, Indiana University}
\affil[3]{MRC Unit for Lifelong Health and Ageing at UCL, University College London}    
\affil[4]{Centre for Medical Image Computing, Department of Computer Science, University College London\vspace{0pt}}

\affil[*]{\texttt{mariana.da\_silva@kcl.ac.uk}\vspace{-15pt}}

\begin{document}
\maketitle

\begin{abstract}
  We present a proof-of-concept, deep learning (DL) based, differentiable biomechanical model of realistic brain deformations. Using prescribed maps of local atrophy and growth as input, the network learns to deform images according to a Neo-Hookean model of tissue deformation. The tool is validated using longitudinal brain atrophy data from the Alzheimer's Disease Neuroimaging Initiative (ADNI) dataset, and we demonstrate that the trained model is capable of rapidly simulating new brain deformations with minimal residuals. This method has the potential to be used in data augmentation or for the exploration of different causal hypotheses reflecting brain growth and atrophy.     
 \end{abstract}

\section{Introduction}

The development of biophysical models of tissue growth and atrophy has historically played a central role in advancing causal understanding of the processes underpinning these phenomena \cite{richman_mechanical_1975, xu_axons_2010, tallinen_growth_2016}. Various biomechanical models of brain deformation have been proposed, typically based on hyperelastic strain models and implemented using finite element methods (FEM) \cite{tallinen_gyrification_2014} or finite difference methods (FDM) \cite{khanal_biophysical_2014,khanal_biophysical_2016}  with prescribed conditions.
However, these approaches are computationally intensive and slow in practice, as they require that the system be solved for each new image or conditions. 

Recently, new techniques have been proposed for deep learning (DL) driven deformable image registration \cite{balakrishnan_voxelmorph_2019, dalca_unsupervised_2018}. Like biomechanical modelling, image registration seeks to deform images through an iterative process incorporating minimisation of a biomechanically-inspired smoothness penalty \cite{robinson_multimodal_2018, jenkinson_fsl_2012}. The VoxelMorph framework \cite{balakrishnan_voxelmorph_2019,dalca_unsupervised_2018}, for example, is trained on pairs of magnetic resonance images (MRI) and optimises a network capable of predicting the displacement field between them. 
At inference time, the method is shown to be orders of magnitude faster than traditional methods, with comparable performance.

Inspired by the VoxelMorph approach to image registration, here we propose an unsupervised deep learning implementation of the longitudinal brain deformation model of Tallinen \textit{et al.} \cite{tallinen_gyrification_2014}. Rather than using this model on a registration setting, we aim to estimate deformations from prescribed atrophy and growth values corresponding to local volume changes, as proposed by Khanal \textit{et al.} \cite{khanal_biophysical_2014,khanal_biophysical_2016}. The framework is tested on modelling patterns of brain atrophy, determined from the longitudinal Alzheimer's Disease Neuroimaging Initiative (ADNI) dataset \cite{petersen_alzheimers_2010}. 

\section{Methods}

\begin{figure}[t]
  \centering
  \includegraphics[width=0.79\textwidth]{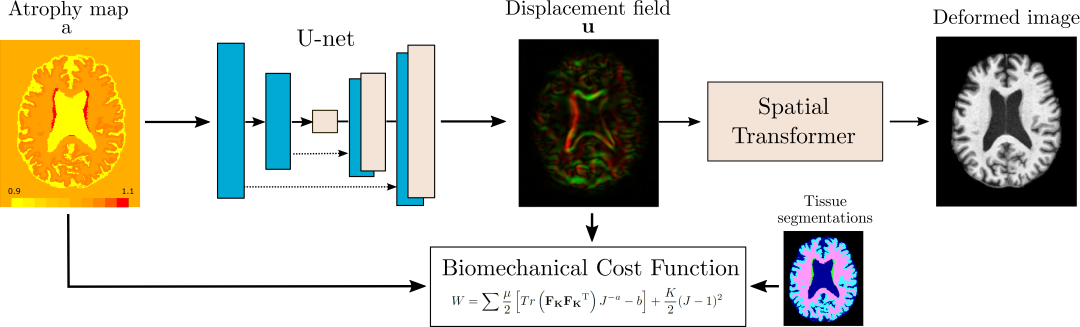}
  \vspace{-4pt}
  \caption{Proposed algorithm: A U-net is optimised with a biomechanical cost function to estimate a displacement field from an atrophy map, which can then be applied to the images using a spatial transformer, in a fully differentiable framework.}
  \label{fig:model}
  \vspace{-10pt}
\end{figure}

\paragraph{Network overview}
We build on the VoxelMorph architecture \cite{balakrishnan_voxelmorph_2019} and extend it specifically for the task of atrophy modelling. We simulate brain deformation using an atrophy map $a$ as input to the model, in the form of a single scalar for each voxel of the brain (Fig. \ref{fig:model}). A U-net is used to estimate a displacement field {\bf{u}} corresponding to the prescribed atrophy. In training, we optimise the network parameters based on a biomechanics-inspired cost function. At inference time, the estimated displacement field is applied to the original MRI grid using a Spatial Transformer, to compute the atrophied image. We implement our framework using the PyTorch library \cite{NEURIPS2019_9015}.

\paragraph{Biomechanical model}

In finite strain theory, a change in the configuration of a body $x$ can be described by a displacement field $\textbf{u}$, such that the new configuration $X$ is given by $X = x + \mathbf{u}$.  We follow the convention used in modelling growth of biological tissues \cite{rodriguez_stress-dependent_1994,young_automatic_2010}, and decompose the total deformation gradient $\mathbf{F} = \nabla \mathbf{u} + \mathbf{I}$ into applied growth $\mathbf{G}$ and elastic deformation $\mathbf{F_K}$, such that $\mathbf{F} = \mathbf{F_K} \cdot \mathbf{G}$. In our model, $G$ is related to the prescribed map $a$ with $\mathbf{G} = (a^{-1/d})\mathbf{I}$, where $d$ is the number of dimensions. We note that $a$ represents relative changes in volume (or area, in 2D); we also assume isotropic growth/atrophy, but extending the model to anisotropic growth would be mathematically trivial. The elastic deformation can be expressed as $\mathbf{F_K} = \mathbf{F} \cdot \mathbf{G^{-1}}$ and is responsible for driving equilibrium. We model the brain tissue as a Neo-Hookean material, and minimise the strain energy density, W, given by:
\begin{equation}
W=\sum{\frac{\mu}{2}\left[Tr\left(\bf{F_K} \bf{F_K}^{\rm{T}}\right) J^{-a}-b\right]+\frac{K}{2}(J-1)^{2}}
\end{equation}

Here, \(J = \rm{det}(\bf{F_K}), \mu \) is the shear modulus and \( K = 100 \mu\) is the bulk modulus. On volumetric deformations, a = \(2/3\) and b = 3, while for 2D experiments we set a = 1 and b = 2. In order to model different tissue characteristics, we define \( \mu = 1 \) for pixels belonging to the gray matter (GM) and white matter (WM), and model the cerebrospinal fluid (CSF) as quasi-free by setting \( \mu = 0.01 \). In order to further enforce biologically meaningful deformations, we add a loss term to encourage zero displacement outside of the skull and set the centre of mass of the brain as fixed point that should not experience deformation. The total cost function to minimise becomes:
\begin{equation}
\mathcal{L} = W + \lambda_1 \sum{\|\bf{u}_{background}\|}^2 + \lambda_2 \|\bf{u}_{center}\|^2,
\end{equation}
where \( \lambda_1, \lambda_2 \) are hyperparameters weighting the contribution of these terms.

\paragraph{Data}

Training and testing of the model was done using 2D axial slices of longitudinal T1-weighted MRI images (4 to 5 time-points per subject) and corresponding tissue segmentations from 354 subjects, acquired as part of the ADNI study \cite{petersen_alzheimers_2010}. The images were brain-extracted, linearly registered to MNI space using FSL FLIRT \cite{jenkinson_fsl_2012} and normalized to [0,1] range using histogram normalisation. 

\paragraph{Training}
 We validate our model by training and testing it using atrophy maps estimated from longitudinal deformable registration, as done by \cite{khanal_biophysical_2016}. For each subject, we use FSL FNIRT \cite{jenkinson_fsl_2012} to register the 1st time-point scan to each subsequent scan, and use the computed Jacobian values as our prescribed atrophy, which we refer to as $a_{FNIRT}$. We obtain a total of 1274 atrophy maps with a variety of levels of atrophy. The dataset was divided into training, validation and testing sets using an 80/10/10 split. The network was trained for 1000 epochs using ADAM as optimizer with learning rate = $10^{-5}$, batch size = 8, $\lambda_1 = 10^{-1}$ and $\lambda_2 = 10^{2}$.
 
\section{Results}

\begin{figure}[t]
  \centering
  \includegraphics[width=\textwidth]{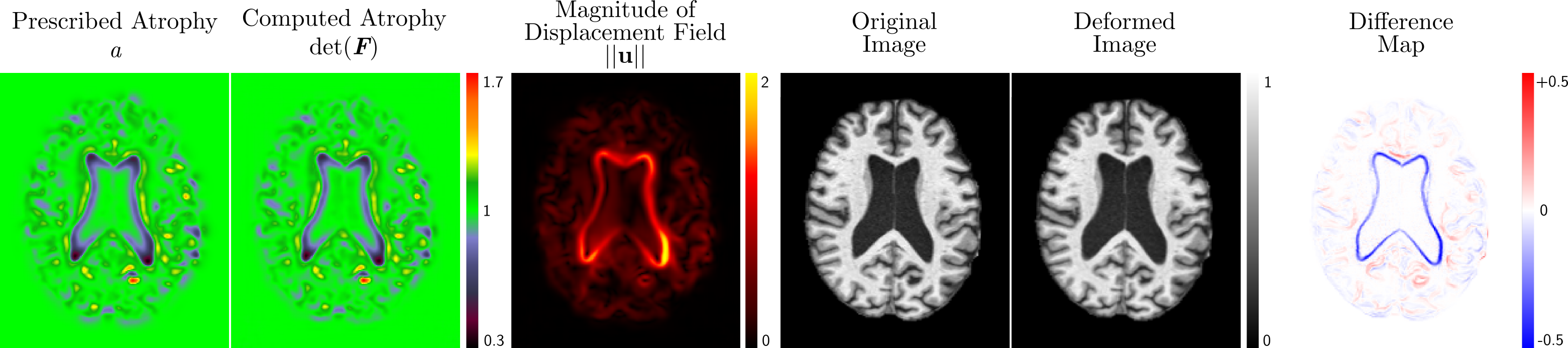}
  \caption{Experimental results on ADNI: Estimating deformations from unseen atrophy maps.}
  \label{fig:results}
  \vspace{-10pt}
\end{figure}

We present the results of the validation experiments on our algorithm using the 36 subjects of the testing dataset. Using the results of FNIRT registration to generate reasonably realistic input atrophy maps to train the model, we show that our network is capable of estimating corresponding displacement fields when provided with unseen atrophy maps (Fig.~\ref{fig:results}). This is translated in small MSE values between input atrophy maps and computed $\rm{det}(\bf{F})$ across all test samples (Table \ref{table1}). To demonstrate the performance of our model, we additionally compute the MSE and Dice scores between the deformed images and the real 2nd time-point images used to estimate the atrophy maps. We report MSE and Dice values comparable to those obtained from the FNIRT registrations.

With the second set of experiments, we show that applying our model by prescribing $a_{FNIRT}$ to the brain tissues only - while allowing the CSF to move freely to compensate this deformation - results in deformation patterns comparable to those obtained using the original atrophy map for all image points. While the MSE between $a_{FNIRT}$ and $\rm{det}(\bf{F})$ increases due to the larger differences in the CSF region, the MSE in the deformed image and Dice coefficient values only show minor changes. This sort of framework can be useful if one wishes to create simple simulations of deformation based on user-defined atrophy values from average volume changes in brain tissue, as explored in \cite{khanal_biophysical_2014}, or for biomechanically-realistic image augmentation.

\begin{table}[h]
  \caption{Comparison between deformed and real images - mean squared error (MSE) and Dice coefficient of different tissues. For our model, we report error between prescribed $a_{FNIRT}$ and the computed atrophy $\rm{det}(\textbf{F)}$. We report average values and corresponding standard deviation.}
  \label{table1}
  \centering
  \small
  \begin{tabular}{lccccccc}
    \toprule
     Method & Input & MSE$_{atrophy}$ & MSE$_{Image}$ & Dice$_{CSF}$ & Dice$_{WM}$ & Dice$_{GM}$ & Dice$_{DGM}$ \\
    \midrule
    FNIRT  &- & - &  \( 9.0\times10^{-4}\) & 0.878 & 0.922 & 0.839 & 0.780 \\
    \vspace{2pt}
    & &  &  ($4.8\times10^{-4}$) & (0.021) & (0.008) & (0.012) & (0.028)\\
    Ours & a$_{FNIRT}$  & \(5.1\times10^{-5}\) & \( 1.0\times10^{-3} \) & 0.877 & 0.921 & 0.837 & 0.775  \\
    \vspace{2pt}
    & & \( (1.8\times10^{-5})\) & \( (5.1\times10^{-4}) \)  & $ (0.021)$ & $(0.009)$ & $(0.013)$ & $(0.032)$\\
    & a$_{FNIRT}$ & \(8.2\times10^{-4}\) & \(1.2\times10^{-3}\) & 0.874 & 0.920 & 0.836 & 0.767 \\
    & brain only & \( (3.9\times10^{-5})\) & \( (6.9\times10^{-4})\) & $ (0.022)$ & $(0.009)$ & $(0.014)$ & $(0.044)$\\
    \bottomrule
  \end{tabular}
  \vspace{-10pt}
\end{table}

\section{Conclusions and Further Work}

In this work we present proof-of-concept for an unsupervised deep learning implementation of a biomechanical model of longitudinal tissue degeneration. The framework was validated with a 2D model of brain atrophy, but can be extended to 3D general models of deformation, as VoxelMorph is a 3D registration network.

We note that one disadvantage of using a DL setting for this type of task is that, unlike in methods such as FEM where the mesh configuration is updated during the iterative optimisation, here we estimate the displacement field from the original image space only. In order to deal with larger or more complex deformation patterns, the current model could be implemented in a recurrent setting. 

The proposed model was shown to be able to rapidly simulate brain degeneration with controlled levels and patterns of atrophy. Not only does it have the potential to be used as a simple tool in data augmentation, but the ability to control different deformations patterns that are biophysically plausible means it could be used as a basis to develop explainable DL frameworks and support causal modelling of mechanisms of development and disease. It is also important to note that this is a fully differentiable model, meaning it can be directly incorporated into more complex deep learning frameworks, or as a drop in replacement to other non-biomechanically-informed deformation models. In future work, we aim to expand this framework with a generative model trained on longitudinal image data in order to predict patterns of atrophy from unseen samples. 

\section*{Broader Impact}

Deep learning has the potential to significantly improve the sensitivity of medical imaging studies, as it has done for natural image analysis. However, one of the biggest pitfalls of DL models is the lack of explainability associated with their decisions. This becomes of particular importance when dealing with complex mechanisms and high-stakes decisions as in the case of medical studies. In recent years, a lot of attention has been given to the development of interpretable DL models, as well as to the importance of causality in the context of medical imaging \cite{castro_causality_2020}.

Our flexible framework has a potential to contribute to the development of more interpretable DL models for neuroimaging studies. By modelling deformation according to biophysical properties while controlling the values and patterns of the changes, our model opens a way to study different causal hypotheses about the mechanisms of development and disease and their translation in medical imaging features. Secondly, modelling well documented and understood mechanisms such as changes in brain shape and volume due to atrophy or growth could allow to disentangle them from more varied or unknown features of brain development or neurological disease, and ultimately improve the study of complex disorders. Finally, the proposed framework also has the potential to be used as a realistic data augmentation tool, which can improve the generalisation ability and robustness of models.



 \begin{ack}

MdS would like to acknowledge funding from the EPSRC Centre for Doctoral Training in Smart Medical Imaging [EP/S022104/1]. The work of ECR and CB was supported by the Academy of Medical Sciences/the British Heart Foundation/the Government Department of Business, Energy and Industrial Strategy/the Wellcome Trust Springboard Award [SBF003/1116] and Wellcome Collaborative Award  [215573/Z/19/Z]. CHS was supported by the Alzheimer's Society Junior Fellowship - AS-JF-17-011

The Titan V GPU used for this research was donated by the NVIDIA Corporation. The data used in this work was funded by the Alzheimer's Disease Neuroimaging Initiative (ADNI) (National Institutes of Health Grant U01 AG024904) and DOD ADNI (Department of Defense award number W81XWH-12-2-0012). 

The Authors declare that there is no conflict of interest.

 \end{ack}


\small

\bibliographystyle{ieeetr}      
\bibliography{Biomechanical.bib}

\end{document}